\documentclass[lettersize,journal]{IEEEtran}
\usepackage{amsmath,amsfonts}
\usepackage{algorithmic}
\usepackage{algorithm}
\usepackage{array}
\usepackage[caption=false,font=normalsize,labelfont=sf,textfont=sf]{subfig}
\usepackage{textcomp}
\usepackage{stfloats}
\usepackage{url}
\usepackage{verbatim}
\usepackage{graphicx}
\usepackage{cite}
\usepackage{multirow} 
\usepackage{booktabs}
\usepackage{rotating}
\usepackage{colortbl}
\newcommand{\ModuleNameA}{cross-modal graph enhancement}
\newcommand{\ModuleA}{CGE}

\newcommand{\ModuleB}{AGR}
\usepackage[table]{xcolor}
\definecolor{row}{HTML}{E9F3FE}
\definecolor{myblue}{HTML}{0155C7}

\begin{document}

\title{DGA-Net: Enhancing SAM with Depth Prompting and Graph-Anchor Guidance for Camouflaged Object Detection}

\author{Yuetong Li, Qing Zhang*, Yilin Zhao, Gongyang Li, Zeming Liu}



\maketitle

\begin{abstract}
To fully exploit depth cues in Camouflaged Object Detection (COD), we present DGA-Net, a specialized framework that adapts the Segment Anything Model (SAM) via a novel ``depth prompting" paradigm. Distinguished from existing approaches that primarily rely on sparse prompts (e.g., points or boxes), our method introduces a holistic mechanism for constructing and propagating dense depth prompts. Specifically, we propose a Cross-modal Graph Enhancement (CGE) module that synthesizes RGB semantics and depth geometric within a heterogeneous graph to form a unified guidance signal. Furthermore, we design an Anchor-Guided Refinement (AGR) module. To counteract the inherent information decay in feature hierarchies, AGR forges a global anchor and establishes direct non-local pathways to broadcast this guidance from deep to shallow layers, ensuring precise and consistent segmentation. Quantitative and qualitative experimental results demonstrate that our proposed  DGA-Net outperforms the state-of-the-art COD methods. 
\end{abstract}


\section{Introduction}
Camouflaged Object Detection (COD) aims to identify and segment objects that are visually concealed within a scene. It has a wide range of real-world applications, including species discovery~\cite{species}, industrial defect detection~\cite{indust}, and medical diagnostics~\cite{PraNet}. 

In recent years, the Segment Anything Model (SAM)~\cite{SAM} has demonstrated strong generalization and zero-shot performance by segmenting arbitrary objects using prompts such as points, boxes, and masks. However, directly applying SAM to COD poses significant challenges. Camouflaged targets typically exhibit ambiguous boundaries, indistinguishable details, and low contrast against their backgrounds, resulting in a substantial mismatch with the natural-image pretraining distribution of SAM. Consequently, this leads to suboptimal segmentation performance.


To address this issue, recent studies have explored various strategies to enhance SAM's adaptability for COD. Existing strategies can be broadly categorized into two types.
The first type focuses on structural adjustments, where lightweight adapters are inserted into the network or the image encoder–decoder is partially redesigned to better tailor SAM to camouflaged object segmentation~\cite{SAM2-UNet,SAM-Adapter}.
The second type extends SAM by introducing auxiliary multi-modal cues, such as depth information or BLIP-generated textual semantics, and typically adopts a dual-branch architecture that processes RGB and the auxiliary modality separately before fusing them at later stages~\cite{DSAM,MM-SAM,SAM-DSA}. For example, Liu \textit{et al.} propose SAM-DSA~\cite{SAM-DSA}, which employs dual-branch RGB–Depth adapters guided by hybrid prompts composed of bounding boxes and depth maps. This design injects geometric cues into SAM while preserving the model’s original sparse prompt interaction paradigm.

\begin{figure*}[!h]
\centering
\includegraphics[width=0.87\textwidth]{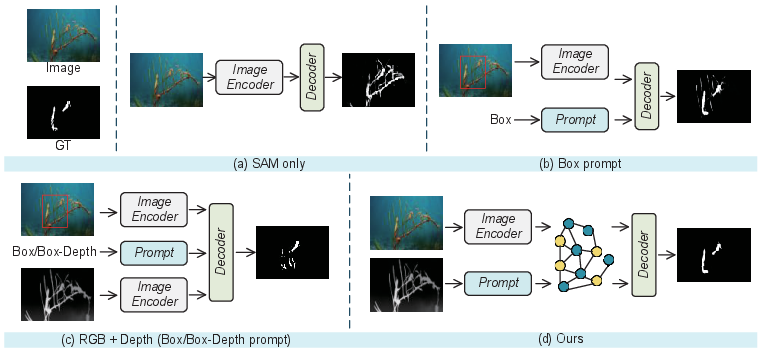} 
\caption{Visualization of different prompting strategies in SAM-based camouflaged object detection. (a) SAM without any prompt.~\cite{SAM2-UNet} (b) Segmentation guided by a box prompt.~\cite{SAM-Adapter} (c) Dual-branch RGB-Depth encoder guided by box or hybrid box-depth prompts.~\cite{DSAM,SAM-DSA} (d) Segmentation guided by a depth-aware geometric prompt and graph-based RGB–Depth fusion (Ours).}
\label{fig1}
\end{figure*}

Although these methods have made progress, they still fail to fully address the challenges of guiding the model in complex scenes with multiple objects, blurred boundaries, and semantic clutter. As shown in Fig.~\ref{fig1}, the plain SAM baseline (a)~\cite{SAM2-UNet} is easily disturbed by background structures under low-contrast conditions, often leading to over-segmentation of the background. Although the box prompt (b)~\cite{SAM-Adapter} provides coarse spatial localization, it also introduces significant background noise, causing the predicted mask to still contain a large amount of irrelevant background regions. Even under an RGB–Depth configuration that follows the dual-branch, box-based prompting paradigm (c)~\cite{DSAM,SAM-DSA}, the results become more compact than the RGB-only baseline, yet the target contours still appear fragmented and discontinuous.

We argue that these problems can be attributed to two key factors. The first, at the modality level, is a representational mismatch between heterogeneous modalities. RGB images provide rich semantic textures, while depth maps offer explicit geometric structures. However, due to their fundamentally different data distributions and physical meanings~\cite{dformerv2}, simple fusion can lead to suboptimal representations where one modality's features overshadow or are misinterpreted by the other. The second is attenuation of prompt guidance within the feature hierarchy. While the structural constraints provided by depth prompts must be transmitted from deep to shallow layers, propagating this signal via standard stepwise pathways inevitably leads to information decay, thereby undermining its ability to strictly regulate fine-grained details.
This raises a key question: how can we synthesize a unified representation from heterogeneous modalities and propagate it throughout the entire hierarchy to ensure global consistency?

To address the above challenges,  we propose DGA-Net, a framework that introduces two synergistic processes to achieve robust camouflaged object detection.
First, the framework introduces a Unified Cross-modal Encoder centered around our Cross-modal Graph Enhancement (CGE). It conceptualizes depth as a dense geometric prompt and models the interaction between PVT-processed RGB features and depth features within a heterogeneous graph. Through bidirectional message passing, the CGE process facilitates mutual learning and calibration, yielding an enhanced, unified representation that is both semantically rich and structurally sound.
Building upon this unified representation, the subsequent Anchor-Guided Refinement (AGR) process then counteracts the hierarchical guidance dilution. It begins by forging a definitive semantic-structural anchor, fusing the enhanced features from CGE with SAM's top-level knowledge. This anchor's guidance is then broadcast to all shallower feature levels via a direct, cross-level information pathway, forcing the entire feature hierarchy to align with a unified global interpretation and ensuring consistent, precise segmentation.

Our contributions are summarized as follows:
\begin{itemize}
\item We propose DGA-Net, a novel framework for COD that enhances SAM with a graph- and anchor-based internal guidance mechanism. Starting from treating depth as a dense prompt, our framework systematically forges a high-fidelity semantic-structural anchor and propagates its guidance across all feature levels.
\item We introduce the CGE process, which resolves the representational mismatch by performing mutual calibration between RGB and Depth modalities within a heterogeneous graph, synthesizing a high-quality, structurally-aware feature source.
\item We further devise the AGR stage, which counteracts the hierarchical guidance dilution by first forging a definitive semantic-structural anchor and then broadcasting its influence throughout the feature hierarchy via a novel cross-level propagation mechanism.
\end{itemize}

\section{Related Work}
\subsection{Camouflaged Object Detection}
With the rapid development of deep learning, the field of COD has made remarkable progress in recent years. Existing methods can generally be divided into three categories: 1) Supplementary information strategy~\cite{FEDER,HGINet,PRBENet}. These approaches incorporate additional cues such as frequency or boundary priors to enhance the discriminative ability of features, thereby alleviating the low contrast between camouflaged objects and their backgrounds. For example, He \textit{et al.}~\cite{FEDER} address the similarity of foreground and background by decomposing the features into different frequency bands using learnable wavelets. Yao \textit{et al.}~\cite{HGINet} propose a region-aware token focusing attention module that enables the model to excavate the distinguishable tokens by employing a dynamic token clustering strategy.
2) Bio-inspired strategy~\cite{SINetV2,Zoomnet}. This category draws inspiration from predator behaviors or human visual perception mechanisms to mimic the process of searching and identifying camouflaged objects. For instance, Fan \textit{et al.}~\cite{SINetV2} simulate the process in which predators search for and recognize camouflaged objects during hunting activities. Pang \textit{et al.}~\cite{Zoomnet} emulates human vision by zooming in and out on imperceptible camouflaged objects and enhance model accuracy through multi-scale fusion.
3) Multi-task joint learning~\cite{CamoFormer,RISNet,CamoDiffusion,RUN,VSCode}. These methods perform progressive reasoning through multi-stage, cascade, or diffusion-based frameworks, leveraging contextual information to improve the localization and segmentation of camouflaged targets. For example, Wang \textit{et al.}~\cite{RISNet} use multi-scale receptive fields to capture feature information of different-sized concealed objects, and propose a large-scale RGB-D agricultural concealed object datasets. Yin \textit{et al.}~\cite{CamoFormer} present a simple masked separable attention that discovers the foreground and background regions by separately computing their attention scores via predicted maps.

Although existing COD methods have made efforts to capture contextual dependencies and relational cues, most of them are limited to local layers or rely on implicit attention schemes.  This limitation hinders their ability to maintain hierarchical consistency across multiple feature levels and to explicitly model cross-modal complementarities. In contrast, our approach formulates the fusion of RGB and depth information as interactions among graph nodes. We construct a heterogeneous graph that spans multiple feature levels and facilitates top-down information propagation through these graph nodes, thereby enabling more comprehensive relation modeling.

\subsection{Segment Anything Model}
In recent years, the SAM has attracted considerable attention as a powerful vision foundation model, primarily due to its strong generalization capability. SAM is designed to segment arbitrary objects using various types of prompts (\textit{e.g.}, points, boxes, masks, and text), enabling its wide applicability across diverse scenarios. However, in the COD domain, targets usually exhibit extremely subtle differences and low contrast against their backgrounds. This characteristic results in a significant semantic gap between COD data and the natural images on which SAM was pretrained. Consequently, directly applying SAM to COD task typically leads to unsatisfactory segmentation performance or even erroneous results, failing to meet practical requirements. 
To address these challenges, recent studies have explored various strategies to enhance the adaptability of SAM in COD tasks. 
Some methods introduce lightweight adapters or redesigned decoders to improve SAM’s feature representation capability~\cite{SAM-Adapter,SAM2-UNet,COD-SAM}.
Others focus on incorporating additional prompting strategies to strengthen the semantic alignment between prompts and visual features~\cite{MM-SAM}. Meanwhile, several works adopt multi-modal inputs to leverage cross-modal cues for improved geometric and semantic perception~\cite{DSAM,SAM-DSA}. For example, Ren \textit{et al.}~\cite{MM-SAM} propose a novel framework to generate multi-modal prompts, thus eliminating the need for manual prompts. Liu \textit{et al.}~\cite{SAM-DSA} augment the association of dual stream embeddings using bidirectional knowledge distillation. 

Differently, we propose a geometric cue framework for COD, which introduces depth information as a dense cue and transforms it into a semantically aware guidance signal after interacting with RGB features, thereby making SAM better adapted to low-contrast and fine-grained camouflage scenes.

\subsection{Graph-based Modeling in Camouflaged Object Detection}
Graph-based reasoning has been widely applied in various computer vision tasks, such as 3D scene understanding and generation~\cite{Octree-Graph,FROSS,3DGraphLLM}, object detection~\cite{QA-FewDet,GraphFPN,DBGL}, and disease detection and gene expression prediction in the medical field~\cite{M2TGLGO,MERGE,GCN-Lung}. 
In the field of COD, several studies have introduced graph reasoning mechanisms to model the spatial dependencies and semantic correlations between foreground and background regions, thereby better capturing boundary details and contextual information of camouflaged targets.
For example, Zhai et al.~\cite{MGL} proposed the MGL framework, which performs multi-level feature relation reasoning through graph-based modeling to enhance the contextual understanding of camouflaged objects. Yao \textit{et al.}~\cite{HGINet} proposed HGINet, which integrates hierarchical graph interaction and dynamic token clustering within a Transformer architecture to enhance semantic context modeling.

Different from existing graph-based COD frameworks that mainly focus on region or boundary reasoning, our method employs graph interaction to achieve modality and hierarchy aware fusion. This approach jointly models the relationships between PVT–Depth and PVT–SAM features, thereby constructing a structure aware cross-modal fusion pipeline for camouflaged object detection.

\begin{figure*}[!htbp]
\centering
\includegraphics[width=1\textwidth]{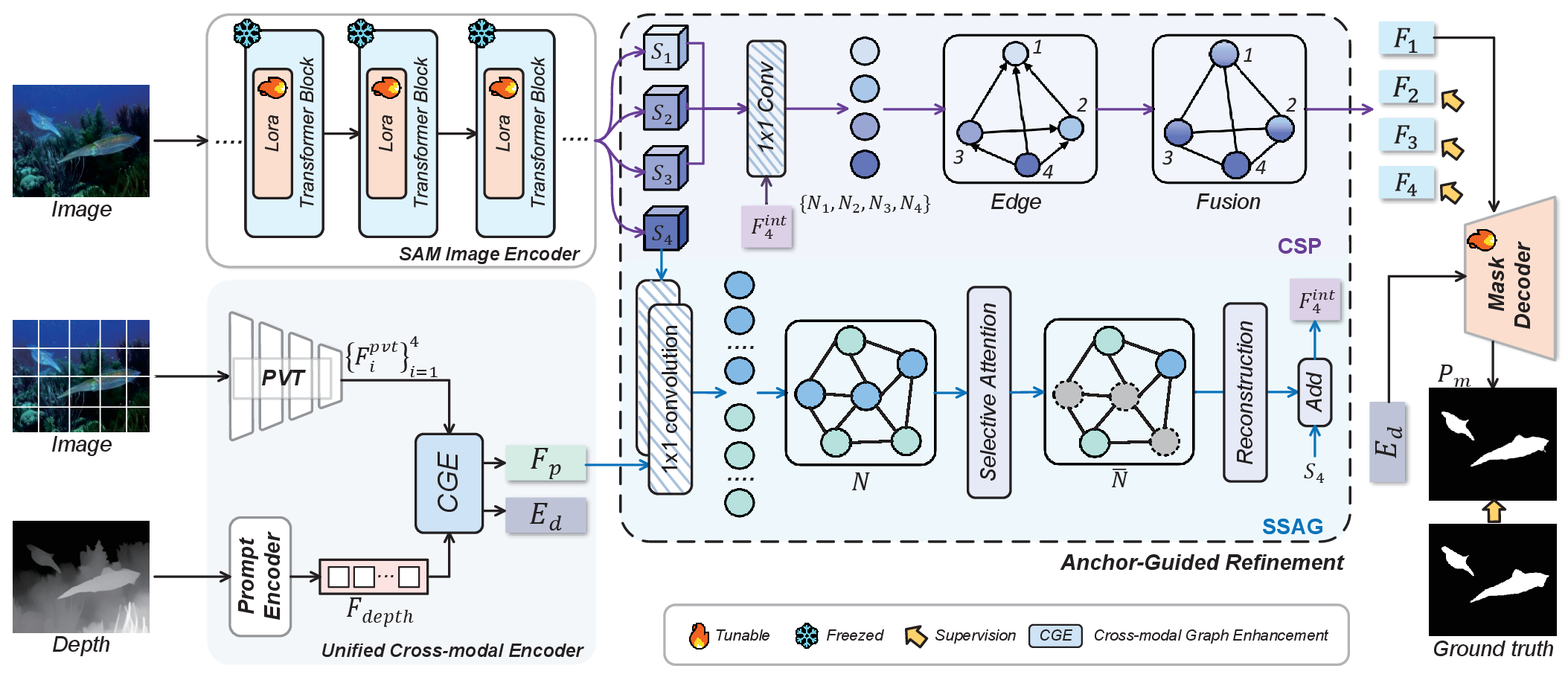} 
\caption{Overview of our proposed DGA-Net, which consists of four main components, i.e., SAM encoder, unified cross-modal encoder, anchor-guided refinement (AGR) and SAM decoder. }
\label{fig:main}
\end{figure*}

\section{Method}
\subsection{Overall}
As illustrated in Fig.~\ref{fig:main}, our framework comprises two parallel encoders: a frozen SAM Encoder (optimized with LoRA) and our Unified Cross-modal Encoder. Crucially, Unified Cross-modal Encoder treat depth information strictly as a dense structural constraint that conditions the pre-trained feature space. Specifically, it processes the RGB image via a PVT and the depth map $D$ with SAM's prompt encoder. At its core is our proposed CGE, which leverages a heterogeneous graph to enable mutual learning and calibration between the texture-sensitive PVT features and the structurally-rich depth features. This produces a semantically-aligned and structurally-aware guidance source. Subsequently, the AGR stage propagates this guidance by forging a global anchor and broadcasting its unified interpretation directly across the entire feature hierarchy $\{S_i\}_{i=1}^4$, effectively counteracting information decay. Finally, the detailed refined feature $F_1$ and an enhanced depth cue $E_d$ feed into the Mask Decoder to generate the highly accurate final prediction $P_m$.

\subsection{Cross-modal Graph Enhancement}

\begin{figure*}[!htbp]
\centering
\includegraphics[width=0.90\textwidth]{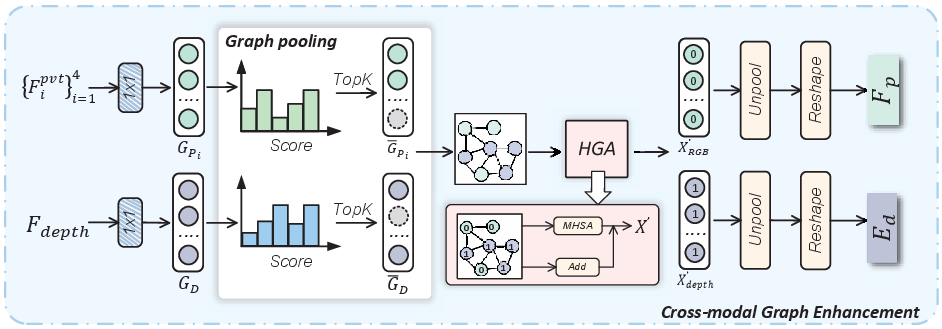} 
\caption{The details of our cross-modal graph enhancement (CGE) module.}
\label{fig:CGE}
\end{figure*}

The distinct characteristics of depth and RGB modalities allow them to complement each other and enhance their respective representations. However, a fundamental challenge in fusing RGB and depth information lies in the modality representation heterogeneity gap: a profound disparity rooted in the inherent limitations of each modality's knowledge. In camouflaged scenarios, RGB representations, despite being semantically rich, exhibit a susceptibility to texture ambiguity, leading to unreliable predictions when object boundaries blend with the background. Conversely, depth representations provide robust geometric structures but are characterized by their semantically agnostic nature, lacking the contextual information to distinguish the target from its surroundings. Straightforward interactions (e.g., concatenation or direct attention) are thus insufficient to bridge this gap. To address this, we propose the \ModuleNameA~(\ModuleA), which employs heterogeneous graph learning to achieve mutual learning and calibration between depth features and PVT-processed RGB features. The PVT branch, being more sensitive to textures and fine-grained cues, complements the structural information from depth, leading to cross-modal enhancement and consistent semantic–structural alignment, as illustrated in Fig.~\ref{fig:CGE}.

Specifically, we first apply $1 \times 1$ convolutions to project multi-level RGB features $\{F_i^{pvt}\}_{i=1}^4$ and the depth feature $F_{depth}$ into respective node sets $\{G_{P_i}\}_{i=1}^4$ and $G_D$. 
Next, to distill the most informative representations, we perform graph pooling on each node set to filter out redundancy while preserving critical cues:
\begin{equation}
 \bar{G}_{P_i} = \mathrm{GraphPooling}(G_{P_i}, r_i), \quad \bar{G}_D = \mathrm{GraphPooling}(G_D, r_d),
\end{equation}
where $r_i$ denotes the pooling ratio for the RGB node sets and $r_d$ denotes the pooling ratio for the depth node sets.

After that, we introduce heterogeneous graph attention (HGA) module to facilitate the mutual calibration and fusion between pooled RGB and depth nodes, obtaining enhanced RGB nodes $X'_{RGB}$ and depth nodes $X'_{depth}$. Next, since the graph pooling reduces the number of nodes, an unpooling operation restores the original number of nodes by placing the enhanced nodes back to their initial indices. 

Finally, the restored node sets are reshaped back into spatial feature maps, yielding the enhanced RGB features $F_{p}$ and depth feature $E_{d}$. This process can be formulated as: 
\begin{equation}
\begin{aligned}
F_{p}=\mathrm{Reshape}(\mathrm{Unpool}(X'_{RGB})), \\ \notag
\quad E_{d}=\mathrm{Reshape}(\mathrm{Unpool}(X'_{depth}))
\end{aligned}
\end{equation}
where $\mathrm{Reshape}(\cdot)$ denotes the operation that transforms a set of nodes back into spatial feature maps, and $\mathrm{Unpool}(\cdot)$ represents the unpooling operation.

\subsubsection{Graph Pooling}

In the graph pooling, we aim to preserve critical nodes while discarding redundant ones. Specifically,  inspired by ~\cite{gao2019graph}, we assign an importance score $s_i$ to each node $x_i$ through a learnable projection function: 

\begin{equation}
s_{i}=f_\theta(x_{i})
\end{equation}
where $f_\theta(\cdot)$ is the learnable projection function. Then, we apply a Top-$K$ strategy to retain the top highest-scoring nodes and discard the remaining redundant ones:
\begin{equation}
 \text{TopK}(s, k) = \{x_i | \text{rank}(s_i) \leq k\}
\end{equation}
where $k$ denotes the number of nodes to preserve, which is determined by the pooling ratio $r$, \textit{i.e.}, $k=r\times N$, where $N$ is the total number of nodes. $\text{TopK}(\cdot)$ represents the node selection operation based on scores, and $\text{rank}(s_i)$ indicates the ranking position of $s_i$ among all nodes. 
It is noted that different pooling ratios are employed at different feature levels. For the RGB node at the $i$-th level, the pooling ratio is set to $r_{i}=[0.2,0.4,0.6,0.8]$,  
allowing the network to gradually reduce redundancy while preserving hierarchical semantics.
And a pooling ratio $r_{d} = 0.5$ is applied to the depth node to reduce redundancy while maintaining structural and geometric information. This process results in the pooled RGB nodes $\{\bar{G}_{P_i}\}_{i=1}^4$ and the pooled depth node $\bar{G}_D$. Our graph pooling evaluates the significance of each node based on its features and relationships within the graph, allowing us to effectively identify and retain the most relevant nodes while filtering out those that contribute less to the camouflage representation

\subsubsection{Heterogeneous Graph Attention}
The HGA module serves as the key component to bridge the heterogeneity gap between modality representations. It is specifically designed to enable a mutual learning process, where the semantically rich but ambiguous RGB features and the structurally robust but semantically agnostic depth features can calibrate and enhance one another. To establish a unified interaction space, all pooled nodes from different scales and modalities are concatenated to form a unified node set $X$: 
\begin{equation}
    X=[\bar{G}_{P_1},\bar{G}_{P_2},\bar{G}_{P_3}, \bar{G}_{P_4}, \bar{G}_D]
\end{equation}
where $[\cdot]$ denotes the concatenation operation. 

Critically, to account for their distinct origins, each node is assigned a type indicator (0 for RGB nodes, 1 for depth nodes). This distinction allows for the application of two distinct linear transformations $\phi_{rgb}$ and $\phi_{depth}$, applied to RGB and depth nodes, respectively. This process is designed to transform the features from their original, modality-specific spaces into a common embedding space. This projection serves to learn a modality-aware transformation that best prepares each feature type for cross-modal interaction, while simultaneously harmonizing their feature dimensions for joint processing. Subsequently, the transformed nodes are processed by a Multi-Head Self-Attention (MHSA) block. Within this block, a synergistic disambiguation is achieved through global message passing: every node can attend to all other nodes, allowing for a comprehensive exchange of contextual information. For instance, this permits an RGB node with an uncertain boundary to be refined by structural evidence from relevant depth nodes. This global information exchange is formulated with a residual connection as follows:
\begin{equation}
 X' = X + \mathrm{MHSA}([\phi_{rgb}(X_{RGB}), \phi_{depth}(X_{depth})])
\end{equation}
where $\phi_{rgb}(\cdot)$ and $\phi_{depth}(\cdot)$ denote the linear transformations applied to RGB and depth nodes, and $\mathrm{MHSA}(\cdot)$ denotes the attention mechanism that drives the cross-modal feature calibration. Finally, the updated node set $X'$ is separated back into its respective RGB $X'_{RGB}$ and depth $X'_{depth}$ components.

\subsection{Anchor-Guided Refinement Module}
Although the CGE module generates the dense source of structurally-aware guidance, effectively utilizing this information within SAM's architecture presents a distinct challenge. Mere injection is insufficient, as the guidance signal is prone to dissipation when traversing the multi feature layers, leading to a loss of constraint integrity. Specifically, this externally-derived knowledge must be deeply integrated with SAM's internal representations and its influence conducted throughout the entire feature hierarchy to ensure globally consistent segmentation. 
To this end, we introduce the anchor-guided refinement (AGR) module, which designed specifically to address this integration and conduction problem through two synergistic stages: (1) The Semantic-Structural Anchor Generation (SSAG) stage, which forges a definitive global anchor by fusing the CGE's enhanced semantics with SAM's top-level knowledge. (2) The Cross-Level Semantic Propagation (CSP) stage, which establishes a direct information pathway to broadcast this anchor's guidance, aligning all shallower features with a unified global interpretation, as illustrated in Fig.~\ref{fig:main}.

The {\ModuleB} consists of two stages: (1) graph projection and Semantic-Structural Anchor Generation (SSAG): the enhanced PVT features $F_{p}$ and the SAM feature $S_4$ are projected into a unified node space, where MHSA selectively integrates the discriminative semantics, yielding the enhanced feature $F^{int}_{4}$. (2) Cross-Level Semantic Propagation (CSP): $F^{int}_{4}$ is combined with SAM’s shallow features $\{S_i\}_{i=1}^3$ to construct a directed hierarchical graph. Through graph-driven top-down semantic propagation, cross-layer collaboration is achieved, ultimately yielding the enhanced multi-scale features $\{F_{i}\}_{i=1}^{4}$.

\subsubsection{Semantic-Structural Anchor Generation }
The SSAG stage forges a global semantic anchor $F_4^{int}$ by synergistically integrating external knowledge with SAM's internal representations. This semantically robust and structurally precise representation then serves as the authoritative guidance for subsequent refinement stages.
Specifically, this integration process involves three sequential operations, including Projection, Selective Attention, and Reconstruction, which work together to integrate the enhanced PVT feature $F_p$ and the SAM feature $S_4$ through graph node interaction.
The projection stage begins by applying a $1\times1$ convolution for embedding alignment, followed by a reshaping operation that converts the spatial features of $F_p$ and $S_4$ into graph node representations within a unified embedding space.
The produced nodes, denoted as $G_{F_p}$ and $G_{S_4}$, are then concatenated to construct a joint node set $N$:
\begin{equation}
N = \left[\,G_{F_p} \,,\, G_{S_4} \,\right]
\end{equation}

In the subsequent Selective Attention block, we aim to refine the joint node set $N$ by focusing the interaction on the most critical features. In camouflaged scenes, the feature maps from both $F_p$ and $S_4$ inevitably contain numerous nodes corresponding to background regions, which can be noisy or even misleading. Including these low-information nodes in the global attention calculation can dilute the focus on meaningful foreground cues. Therefore, a learnable graph pooling operation with a ratio of $r=0.7$ is applied to retain the top 70\% highest-scoring nodes from the constructed graph $N$, and their indices are stored for subsequent unpooling. Afterwards, the MHSA is performed on the denoised node set to facilitate a deep, global interaction. This process enables a synergistic fusion of the features' respective strengths: the structural fidelity from 
$F_p$ is integrated with the generalized semantic knowledge from $S_4$.

\begin{equation}
\bar{N} = \mathrm{MHSA}(\mathrm{GraphPooling} (N))
\end{equation}
where $\bar{N}$ denotes the refined node set after graph pooling and MHSA.

Finally, the reconstruction stage maps the refined node representations in {N} back into a spatial feature map. First, an unpool operation utilizes the stored pooling indices to restore the condensed node set {N} to its original full-sized set, $S_{{unpool}}$. 

From this set, we extract the subset of nodes, $S_{{feat}}$, corresponding to the original spatial locations of the $S_4$ feature. This subset, representing the learned enhancement for SAM's top-level feature, is reshaped back into a 2D feature map. This map is added to the original $S_4$ feature via a residual connection, yielding the final global semantic anchor $F_4^{int}$.

\begin{equation}
F^{int}_{4}=\text{Reshape}(S_{feat})+S_{4}
\end{equation}

\subsubsection{Cross-Level Semantic Propagation }
To mitigate the attenuation of prompt typically associated with sequential layer-wise transmission, we introduce the Cross-Level Semantic Propagation stage. Its core task is to propagate the anchor's guidance from the top level to all shallower feature maps.
To achieve this, CSP establishes a directed hierarchical graph where the global anchor acts as the root node. From this root, we construct direct, non-local connections to all shallower feature nodes $\{S_1, S_2, S_3\}$. Through these connections, the anchor's guidance is broadcast simultaneously, allowing each shallow node to aggregate these top-down messages with its own local features. This process ensures the entire feature hierarchy is recalibrated according to a unified and coherent global plan, ultimately yielding a new, internally coherent set of enhanced features $\{F_{i}\}_{i=1}^{4}$.

Specifically, we utilize $F^{int}_{4}$ along with the shallow-level SAM features $S_{1}, S_{2},$ and $S_{3}$ as inputs. We begin by transforming these spatial features into node representations, which serve as the nodes $\{N_1, N_2, N_3, N_4\}$ of our directed hierarchical graph. This projection is achieved via independent $1 \times 1$ convolutions for each feature level.
To facilitate top-down semantic propagation, we construct the directed edges of our hierarchical graph, which represent the direct, non-local information pathways. Guidance from higher-level nodes is transmitted to lower-level nodes via convolution and upsampling operations:
\begin{equation}
\left\{
\begin{aligned}
E_{4\to3} &= \mathrm{Up} \big(\mathrm{Conv}_{4\to3}(N_4)\big) \\
E_{4\to2} &= \mathrm{Up}^2\!\big(\mathrm{Conv}_{4\to2}(N_4)\big)\\
E_{4\to1} &= \mathrm{Up}^3\!\big(\mathrm{Conv}_{4\to1}(N_4)\big)
\end{aligned}
\right.
\end{equation}
where $\mathrm{Conv}(\cdot)$ denotes the $1\times1$ convolution, $Up^k(\cdot)$ represents bilinear upsampling performed $k$ times, and $E_{t_1 \to t_2}$ denotes the message features transmitted from node $N_{t_1}$ to node $N_{t_2}$.

Similarly, the information from middle-level nodes is also propagated along directed edges to low-level nodes:
\begin{equation}
\left\{
\begin{aligned}
E_{3\to2} &= Up \big(\mathrm{Conv}_{3\to2}(N_3)\big)\\
E_{3\to1} &= Up^2\big(\mathrm{Conv}_{3\to1}(N_3)\big)\\
E_{2\to1} &= Up \big(\mathrm{Conv}_{2\to1}(N_2)\big)
\end{aligned}
\right.
\end{equation}

During the node information fusion, each shallow node $N_{i}$ (for $i \in \{1, 2, 3\}$) concatenates its own feature together with the incoming messages from higher-level nodes along the channel dimension. These aggregated features are then processed by corresponding fusion function $\mathcal{F}(\cdot)$ to generate the semantically enriched fused features ${F_1}$, ${F_2}$ and ${F_3}$. As the source of the guidance, the top-level feature $F_4$ is set directly to its original input $F_4^{int}$.
\begin{equation}
\left\{
\begin{aligned}
F_1 &= \mathcal{F}_1\big[(N_1, E_{2\to1}, E_{3\to1}, E_{4\to1})]\\
F_2 &= \mathcal{F}_2\big[(N_2, E_{3\to2}, E_{4\to2})] \\
F_3 &= \mathcal{F}_3\big[(N_3, E_{4\to3})] \\
F_4 &= F_4^{int}
\end{aligned}
\right.
\end{equation}
where $\mathcal{F}_3$, $\mathcal{F}_2$, and $\mathcal{F}_1$ represent the fusion functions for each node, implemented as two consecutive convolutional layers.

\subsection{Loss Function}
During training, we adopt a hybrid loss function composed of a Binary Cross-Entropy (BCE) loss $\mathcal{L}_{\text{bce}}$~\cite{f3net} and an Intersection-over-Union (IoU) loss $\mathcal{L}_{\text{iou}}$~\cite{f3net} to measure the discrepancy between the ground truth and two sets of predictions. The first set is the final prediction $P_{m}$, which is generated by the SAM Mask Decoder. The second set consists of auxiliary side predictions $\{P_{i}\}_{i=2}^{4}$ that are generated from the features $\{F_{i}\}_{i=2}^{4}$ via $1\times1$ convolutions followed by a Sigmoid activation. The overall loss of our network is defined as:
\begin{equation}
\mathcal{L} = \mathcal{L}_{\text{bce}}(G,P_{m})+\mathcal{L}_{\text{iou}}(G,P_{m})
+ \sum_{i=2}^{4}\Big(\mathcal{L}_{\text{bce}}(G,P_i)+\mathcal{L}_{\text{iou}}(G,P_i)\Big)
\end{equation}
where $G$ is the ground truth.

\section{Experiments}
\subsection{Experimental Settings}
We implement our network using the PyTorch~\cite{pytorch} framework, and all experiments are conducted on a single NVIDIA GTX 3090 GPU. During training and testing, input images are uniformly resized to a spatial resolution of $512\times512$. To enhance data diversity, we apply data augmentation, including random rotation, cropping, and color jittering. For optimization, we employ the Adam optimizer with an initial learning rate of 5e-5, a batch size of 4, and train the network for a total of 60 epochs.
\subsubsection{Datasets}
Three widely used COD benchmarks, namely CAMO~\cite{CAMO}, COD10K~\cite{SINet}, and NC4K~\cite{SLSR}, are adopted in our experiments. CAMO~\cite{CAMO} contains 1,250 camouflaged images, among which 1,000 samples are selected for training while the remaining 250 images are for evaluation. COD10K~\cite{SINet} is the largest COD dataset, comprising a total of 5,066 camouflaged images. Following the common practice, 3,040 camouflaged images are used for training and the remaining 2,026 for testing. NC4K~\cite{SLSR} offers 4,121 Internet-collected images and serves exclusively as a test set to assess the generalization ability of different COD models. 

\subsubsection{Evaluation Metrics}
To comprehensively evaluate the performance of our model, we adopt four commonly used metrics for camouflaged object detection: mean absolute error ($\mathcal{M}$)~\cite{mae}, weighted F-measure ($F_{\beta}^{\omega}$)~\cite{FT}, E-measure ($E_{\phi}^{m}$)~\cite{Emeasure} and S-measure ($S_{m}$)~\cite{sm}. Specifically, $\mathcal{M}$ is a pixel-level metric that measures the average absolute difference between the predicted mask and the ground truth. The weighted F-measure $F_{\beta}^{\omega}$ performs position-sensitive weighting on precision and recall, placing more emphasis on object boundary pixels and thus providing a more balanced evaluation. E-measure $E_{\phi}^{m}$ integrates both local and global information to describe the overall similarity between the prediction and ground truth. $S_{m}$ evaluates the spatial structural consistency between the predicted mask and ground truth from both region-aware and object-aware perspectives. In general, higher values of $F_{\beta}^{\omega}$, $E_{\phi}^{m}$ and $S_{m}$ indicate better model performance, while a lower $\mathcal{M}$ signifies improved performance.

\subsection{Comparison with State-of-the-art Methods}

We compare our netowrk with several representative methods, including MGL~\cite{MGL}, BGNet~\cite{BGNet}, ZoomNet~\cite{Zoomnet}, SINetV2~\cite{SINetV2}, FEDER~\cite{FEDER}, CamoFormer~\cite{CamoFormer}, RISNet~\cite{RISNet}, VSCode~\cite{VSCode}, HGINet~\cite{HGINet}, CamoDiffusion~\cite{CamoDiffusion}, RUN~\cite{RUN}, PRBE-Net~\cite{PRBENet}, SAM~\cite{SAM}, SAM-Adapter~\cite{SAM-Adapter}, SAM2-UNet~\cite{SAM2-UNet}, DSAM~\cite{DSAM}, COD-SAM~\cite{COD-SAM}, VL-SAM\cite{MM-SAM}, SAM-DSA~\cite{SAM-DSA}. For a fair comparison, we use the maps provided by the authors or run the released codes with the recommended settings.

\subsubsection{Quantitative Comparison}
Table~\ref{tab:tb1} summarizes the quantitative comparison between our proposed DGA-Net and other competing methods. It can be observed that our method demonstrates competitive performance across all datasets, achieving either the best or second-best results in all evaluation metrics. 
On the COD10K dataset, our model achieves the most competitive results. Compared with the recently published COD-SAM~\cite{COD-SAM} and SAM-DSA~\cite{SAM-DSA} models, our method consistently outperforms them in terms of $S_m$, $F_{\beta}^{\omega}$, and $\mathcal{M}$. We attribute this improvement to the proposed graph-guided depth modeling strategy, which treats the depth modality as a dense geometric prompt and performs modality interaction in the graph domain, effectively mitigating the bias introduced by traditional prompting mechanisms and the inconsistencies stemming from cross-modal discrepancies. As a result, we achieve more accurate and consistent camouflaged object detection.

\begin{table*}[!htbp]
\scriptsize
\centering
 \renewcommand{\arraystretch}{1.3}
\renewcommand{\tabcolsep}{0.9mm}
\caption{Quantitative comparison with state-of-the-art methods for COD on three benchmarks using four evaluation metrics. "$\uparrow$" / "$\downarrow$" indicates that higher/lower is better. Top two results are highlighted in \textcolor{red}{red} and \textcolor{myblue}{blue}.}
	\label{tab:tb1}
	\scalebox{0.9}{
\begin{tabular}{c|c|c|c|cccc|cccc|cccc} 
   \toprule
    \multirow{2}*{\textbf{Method}}& \multirow{2}*{\textbf{Publication}}& \multirow{2}*{\textbf{Prompt}}& \multirow{2}*{\textbf{Backbone}} &\multicolumn{4}{c}{\textbf{CAMO(250 images)}} &  \multicolumn{4}{c}{\textbf{COD10K(2026 images)}}& \multicolumn{4}{c}{\textbf{NC4K(4121 images)}}\\
    &&&&$S_m\uparrow$&$F^{\omega}_{\beta}\uparrow$&$\mathcal{M}\downarrow$&$E_{\phi}^m\uparrow$&$S_m\uparrow$&$F^{\omega}_{\beta}\uparrow$&$\mathcal{M}\downarrow$&$E_{\phi}^m\uparrow$&$S_m\uparrow$&$F^{\omega}_{\beta}\uparrow$&$\mathcal{M}\downarrow$&$E_{\phi}^m\uparrow$\\
 \midrule
\multicolumn{16}{c}{Non-Large Model Methods}\\
\midrule
MGL & 21CVPR & None & ResNet & 0.775 & 0.673  & 0.088 & 0.847 & 0.814 & 0.666 & 0.035 & 0.865  & 0.833 & 0.740 & 0.052 & 0.867 \\
BGNet & 22IJCAI &  None & Res2Net & 0.812 & 0.749 & 0.073 & 0.870 & 0.831 & 0.722 & 0.033 & 0.901 & 0.851 & 0.788 & 0.044 & 0.907 \\
ZoomNet & 22CVPR &  None & ResNet & 0.820 & 0.752 & 0.066 & 0.877 & 0.838 & 0.729 & 0.029 & 0.888 & 0.853 & 0.784 & 0.043 & 0.896 \\
SINetV2 & 22TPAMI &  None & Res2Net & 0.820 & 0.743 & 0.071 & 0.882 & 0.815 & 0.680 & 0.037 & 0.887 & 0.847 & 0.770 & 0.048 & 0.903 \\
FEDER & 23CVPR &  None & ResNet & 0.836 & 0.807 & 0.066 & 0.897 & 0.844 & 0.748 & 0.029 & 0.911 & 0.862 & 0.824 & 0.042 & 0.913 \\
CamoFormer & 24TPAMI &  None & PVT & 0.872 & 0.831 & 0.046 & 0.929 & 0.869 & 0.786 & 0.023 & 0.932 & 0.892 & 0.847 & 0.030 & 0.939 \\
RISNet & 24CVPR & Depth & ResNet & 0.870 & 0.827 & 0.050 & 0.922 & 0.873 & 0.799 & 0.025 & 0.931 & 0.882 & 0.834 & 0.037 & 0.926 \\
VSCode & 24CVPR &  2D Prompt & Swin & 0.873 & 0.820 & 0.046 & 0.925 & 0.869 & 0.780 & 0.025 & 0.931 & 0.882 & 0.841 & 0.032 & 0.935 \\
HGINet & 24TIP &  None & ViT & 0.874 & 0.848 & \textcolor{myblue}{0.041} & 0.937 & 0.882 & 0.821 & \textcolor{myblue}{0.019} & \textcolor{myblue}{0.949} & 0.894 & 0.865 & \textcolor{myblue}{0.027} &\textcolor{myblue}{0.947} \\
CamoDiffusion & 25TPAMI &  None & PVT, UNet & 0.878 & 0.853 & 0.042 & 0.940 & 0.881 & 0.814 & 0.020 & 0.944 & 0.893 & 0.859 & 0.029 & 0.942 \\
RUN & 25ICML & None & PVT & 0.877 & \textcolor{myblue}{0.861} & 0.045 & 0.934 & 0.878 & 0.810 & 0.021 & 0.941 & 0.892 & \textcolor{myblue}{0.868} & 0.030 & 0.940 \\
PRBE-Net & 25TMM &  None & PVT & 0.876 & 0.837 & 0.045 & 0.928 & 0.867 & 0.793 & 0.021 & 0.932 & 0.887 & 0.845 & 0.031 & 0.931 \\
\midrule
\multicolumn{16}{c}{Large-Model Methods}\\
\midrule
SAM & 23ICCV & None & SAM & 0.684 & 0.606 & 0.132 & 0.687 & 0.783 & 0.701& 0.049 & 0.798 & 0.767 & 0.696 & 0.078 & 0.776\\
SAM-Adapter & 23ICCVW  & Box & SAM & 0.847 & 0.765 & 0.070 & 0.873 & 0.883 & 0.801 & 0.025 & 0.918 & - & - & - & - \\
SAM2-UNet &24Arxiv & None & SAM& \textcolor{myblue}{0.884} & \textcolor{myblue}{0.861} & 0.042 & 0.932 & 0.880 & 0.789 & 0.021 & 0.936 & \textcolor{myblue}{0.901} & 0.863 & 0.029 & 0.941 \\
DSAM & 24MM & Box &SAM,PVT & 0.832 & 0.794 & 0.061 & 0.913 & 0.846 & 0.760 & 0.033 & 0.931 & 0.871 & 0.826 & 0.040 & 0.932 \\
COD-SAM & 25PR & Corner & SAM & 0.870& 0.796 & 0.055 & 0.906 & \textcolor{myblue}{0.899} & \textcolor{myblue}{0.832} &0.021 & 0.941&-&-&-&-\\
VL-SAM & 25ICCV & Text,Vision & SAM & 0.863 & 0.782 & 0.059 & 0.901 & 0.896 & 0.808 & 0.023 & 0.907 & -& -& -& -\\
SAM-DSA & 25ICCV & Box,Depth & SAM,PVT & 0.875 & 0.849 & 0.044 & \textcolor{red}{0.952} & 0.887 & 0.827 & 0.022 & 0.948 & 0.896 & 0.866 & 0.029 & \textcolor{red}{0.959}\\
\rowcolor{row}Ours & - & Depth & SAM,PVT & \textcolor{red}{0.906} & \textcolor{red}{0.877} & \textcolor{red}{0.033} & \textcolor{myblue}{0.946} & \textcolor{red}{0.903} & \textcolor{red}{0.847} & \textcolor{red}{0.018} & \textcolor{red}{0.951} & \textcolor{red}{0.911} & \textcolor{red}{0.878} & \textcolor{red}{0.026} & \textcolor{myblue}{0.947} \\
    \bottomrule
\end{tabular}
}
\end{table*}


\subsubsection{Qualitative Evaluation}
\begin{figure*}[!htbp]
\centering
\includegraphics[width=0.83\textwidth]{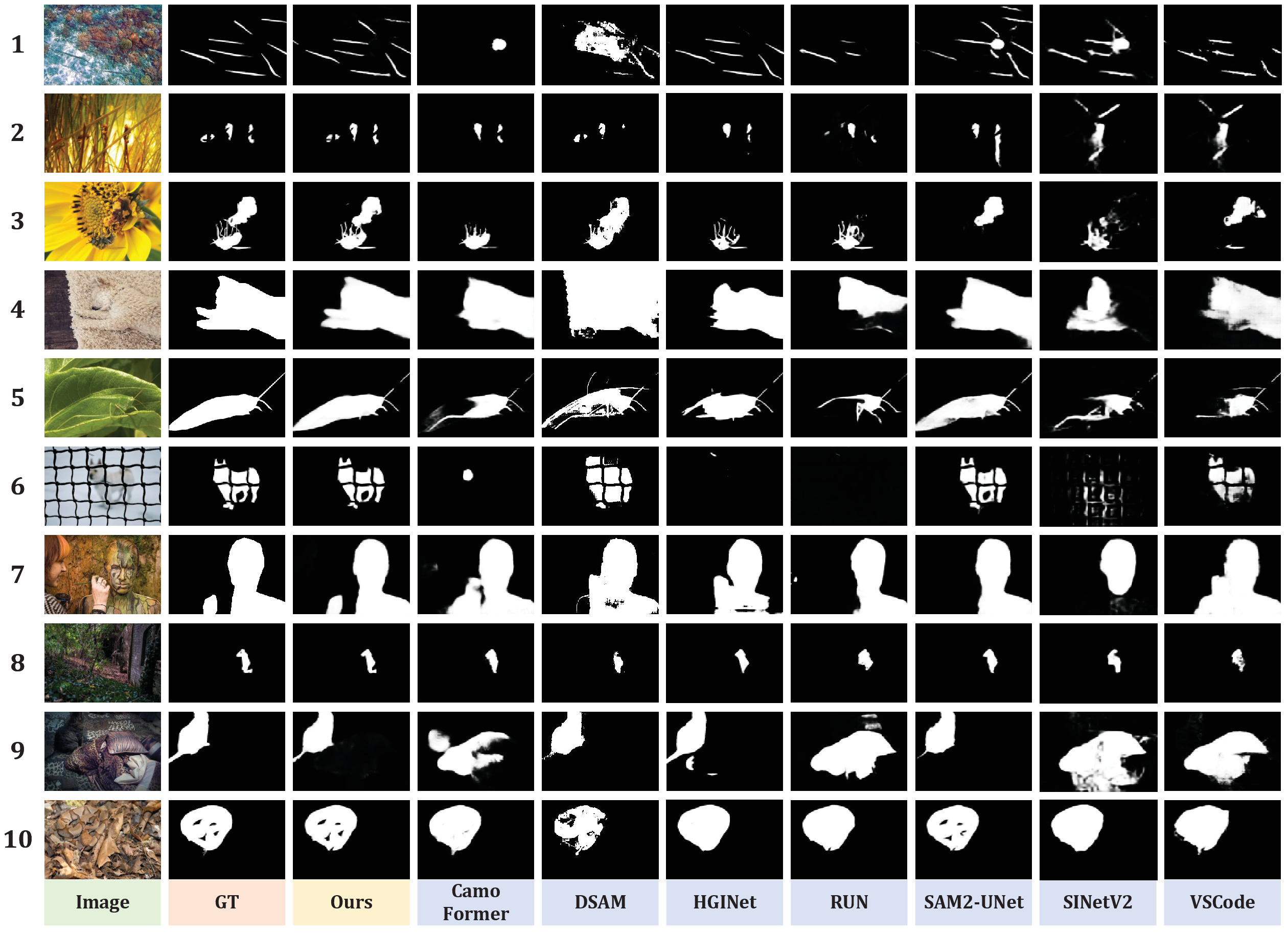} 
\caption{Visual comparisons of some recent COD methods and our proposed network.}
\label{fig:vision}
\end{figure*}

Fig.~\ref{fig:vision} presents a visual comparison between our proposed DGA-Net and other state-of-the-art  methods under diverse and complex scenes. 
These scenes include multiple objects (1st–3rd rows), edge similarity (4th–5th rows), occluded scenes (6th–7th rows), small objects (8th row), and semantic clutter (9th–10th rows). Under such extreme conditions, existing methods often suffer from missing regions, blurred boundaries, incomplete predictions or background misidentification. In contrast, our method effectively maintains structural consistency and boundary integrity across these challenging scenarios. Even in highly confusing or low-contrast regions, our network is able to  accurately localize the camouflaged targets, demonstrating its superior segmentation performance.

\subsection{Ablation Studies}
In this section, we conduct ablation studies on different key components of our model to investigate their influence on the overall performance. 

\subsubsection{Effectiveness of Key Components}
To evaluate the contributions of the two main components, CGE and AGR, we first construct a baseline model by removing these components, referred as ``B". We then progressively add CGE and AGR to the baseline to assess their individual impacts on model performance. In addition, we construct a depth-free variant (denoted as ``w/o Depth") that removes all depth inputs and their associated feature interactions. This allows to evaluate the contribution of depth cues to model performance.

As shown in Table~\ref{tab:ablation1}, by comparing ``B+CGE”, ``B+AGR”, and the complete version of our network (“Ours”), we observe that the introduction of CGE significantly improves performance, validating its effectiveness. This improvement mainly stems from CGE's use of heterogeneous graph modeling to align cross-modal features and facilitate semantic interaction between RGB and depth modalities, thereby producing more consistent fused representations.
When the AGR module is added, the performance is further enhanced, suggesting that AGR can effectively leverage graph-structured information to refine high-level semantics, enhance structural details, and improve boundary localization.
Moreover, the performance of ``w/o Depth” decreases, indicating that depth information provides valuable geometric priors for the model. This further implies that the performance gain by AGR is not solely due to its architectural design but is also enhanced by depth-guided semantic modulation.

\begin{figure}[!htbp]
\centering
\includegraphics[width=0.47\textwidth]{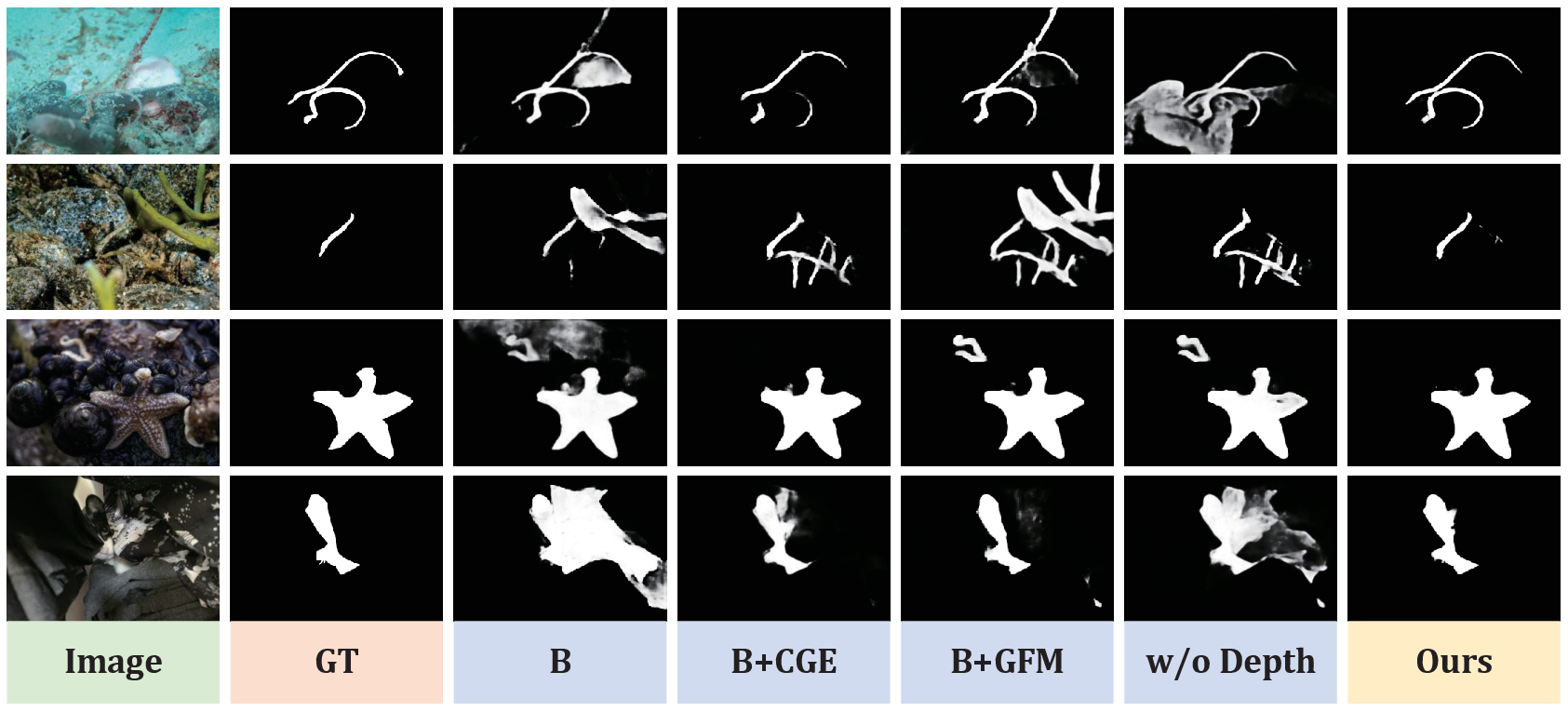} 
\caption{Visualization results of the ablation study on different key components.}
\label{fig:ab1}
\end{figure}

Importantly, when both CGE and AGR are enabled simultaneously (``Ours"), the model achieves the best performance across all metrics. This confirms the complementarity of the two modules: CGE focuses on modeling global cross-modal semantic relationships, while AGR further refines spatial structures and suppresses irrelevant responses, enabling more accurate and coherent perception of camouflaged objects.
In addition, Fig.~\ref{fig:ab1} provides several visual comparisons, from which it can be observed that the segmentation results become more accurate as CGE, AGR and depth information are progressively introduced into the model.

\begin{table}[h!]
\centering
\caption{Ablation Study of our Key Components}
\scalebox{0.75}{
\begin{tabular}{l|ccc|ccc|ccc} 
   \toprule
    \multirow{2}*{\textbf{Method}} &\multicolumn{3}{c|}{\textbf{CAMO}} &  \multicolumn{3}{c|}{\textbf{COD10K}}& \multicolumn{3}{c}{\textbf{NC4K}}\\
\cmidrule[0.05em](lr){2-10}
&$S_m\uparrow$&$F^{\omega}_{\beta}\uparrow$&$\mathcal{M}\downarrow$&
$S_m\uparrow$&$F^{\omega}_{\beta}\uparrow$&$\mathcal{M}\downarrow$&
$S_m\uparrow$&$F^{\omega}_{\beta}\uparrow$&$\mathcal{M}\downarrow$\\
\midrule
B & 0.877 & 0.832 & 0.044 & 0.876 & 0.793 & 0.024 & 0.897 & 0.849 & 0.033\\
B+CGE & 0.886 & 0.835 & 0.042 & 0.878 & 0.799 & 0.023 & 0.897 & 0.850 & 0.032\\ 
B+AGR & 0.887 & 0.840 & 0.043 & 0.879 & 0.793 & 0.024 & 0.902 & 0.854 & 0.030\\
w/o Depth& 0.888 & 0.840 & 0.041 & 0.875 & 0.788 & 0.023 & 0.897 & 0.845 & 0.032\\
\midrule
\rowcolor{row}Ours & \textbf{0.906} & \textbf{0.877} & \textbf{0.033} & \textbf{0.903} & \textbf{0.847} & \textbf{0.018} & \textbf{0.911} & \textbf{0.878} & \textbf{0.026} \\
\bottomrule
\end{tabular}
}
\label{tab:ablation1}
\end{table}

\subsubsection{Effectiveness of the CGE}
1) Fusion Strategy: To validate the effectiveness of the fusion strategy within the CGE module, we design three variants: ``w SF", which replaces the graph-based fusion with simple feature concatenation, ``w UG", which constructs a single homogeneous graph by node type insensitivity, and ``w/o GP\&UP”, which removes all graph pooling and unpooling operations. As shown in Table~\ref{tab:ablation2}, all variants lead to performance degradation, indicating that our heterogeneous graph-based fusion strategy is more effective in capturing semantic correlations and structural dependencies between RGB and depth modalities, thereby producing more consistent and discriminative cross-modal representations. Comparing ``w/o GP\&UP" with ``Ours” in Table~\ref{tab:ablation2}, we observe that the completely removal of graph pooling leads to a notable performance drop. This decline is likely due to the disruption of hierarchical node compression, which weakens the graph structure and leads to the propagation of redundant or irrelevant nodes in subsequent layers, thereby impairing cross-scale semantic alignment.

\begin{table}[h!]
\centering
\caption{Ablation Study of the CGE}
\scalebox{0.75}{
\begin{tabular}{l|ccc|ccc|ccc} 
   \toprule
    \multirow{2}*{\textbf{Method}} &\multicolumn{3}{c|}{\textbf{CAMO}} &  \multicolumn{3}{c|}{\textbf{COD10K}}& \multicolumn{3}{c}{\textbf{NC4K}}\\
\cmidrule[0.05em](lr){2-10}
&$S_m\uparrow$&$F^{\omega}_{\beta}\uparrow$&$\mathcal{M}\downarrow$&
$S_m\uparrow$&$F^{\omega}_{\beta}\uparrow$&$\mathcal{M}\downarrow$&
$S_m\uparrow$&$F^{\omega}_{\beta}\uparrow$&$\mathcal{M}\downarrow$\\
\midrule
w/ SF & 0.887 & 0.840 & 0.043 & 0.886 & 0.806 & 0.022 & 0.904 & 0.855 & 0.030\\
w/ UG & 0.890 & 0.847 & 0.041 & 0.887 & 0.814 & 0.020 & 0.906 & 0.864 & 0.028 \\ 
w/o GP\&UP & 0.893 & 0.847 & 0.040 & 0.887 & 0.812 & 0.021 & 0.905 & 0.862 & 0.029\\
\midrule 
w/o HGA & 0.889 & 0.845 & 0.038 & 0.885 & 0.806 & 0.021 & 0.902 & 0.860 & 0.029\\
$\text{HGA}_{n=3}$ & 0.893 & 0.849 & 0.038 & 0.886 & 0.809 & 0.021 & 0.905 & 0.859 & 0.029\\
$\text{HGA}_{n=2}$ & 0.892 & 0.850 & 0.037 & 0.888 & 0.816 & 0.021 & 0.907 & 0.865& 0.028\\
\midrule
\rowcolor{row}Ours & \textbf{0.906} &\textbf{ 0.877} & \textbf{0.033} & \textbf{0.903} & \textbf{0.847} & \textbf{0.018} & \textbf{0.911} & \textbf{0.878} & \textbf{0.026} \\
\bottomrule
\end{tabular}
}
\label{tab:ablation2}
\end{table}

2) HGA: To further investigate the impact of graph reasoning in the CGE module, we construct three variants by altering the number of heterogeneous graph attention (HGA): ``w/o HGA", which removes the HGA, and ``HGA$_{n=3}$" and ``HGA$_{n=2}$", which stack three and two HGAs, respectively. 
As presented in Table~\ref{tab:ablation2}, removing HGA results in noticeable performance degradation, confirming that HGA plays a crucial role in refining structural features and modeling semantic dependencies between RGB and depth modalities. Moreover, compared with our network that using a HGA, increasing the number of HGA does not lead to further improvements. This is likely due to redundant information propagation introduced by deeper graph reasoning. These results indicate that appropriate  graph reasoning is sufficient to model cross-modal dependencies, whereas excessive stacking is unnecessary. 
\begin{table}[h!]
\setlength{\tabcolsep}{4pt}
\caption{Ablation study on the node reduction ratios of RGB features $r_{i}$ in the CGE module.}
\centering
\scalebox{0.75}{
\begin{tabular}{c|ccc|ccc|ccc} 
   \toprule
    \multirow{2}*{$r_{i}$} &\multicolumn{3}{c|}{\textbf{CAMO}} &  \multicolumn{3}{c|}{\textbf{COD10K}}& \multicolumn{3}{c}{\textbf{NC4K}}\\
\cmidrule[0.05em](lr){2-10}
&$S_m\uparrow$&$F^{\omega}_{\beta}\uparrow$&$\mathcal{M}\downarrow$&
$S_m\uparrow$&$F^{\omega}_{\beta}\uparrow$&$\mathcal{M}\downarrow$&
$S_m\uparrow$&$F^{\omega}_{\beta}\uparrow$&$\mathcal{M}\downarrow$\\
\midrule
$[0.2,0.2,0.2,0.2]$&0.893 & 0.847 & 0.040 & 0.887 & 0.812 & 0.021 & 0.905 & 0.862 & 0.029 \\
$[0.5,0.5,0.5,0.5]$ &0.894 & 0.851 & 0.039 & 0.887 & 0.815 & 0.020 & 0.905 & 0.864 & 0.028\\
$[0.8,0.6,0.4,0.2]$ &0.895 & 0.853 & 0.039 & 0.885 & 0.810 & 0.021 & 0.906 & 0.865 & 0.028 \\
$[0.8,0.8,0.8,0.8]$ & 0.895 & 0.852 & 0.038 & 0.883 & 0.812 & 0.021 & 0.901 & 0.859 & 0.028 \\
\midrule
\rowcolor{row}$[0.2,0.4,0.6,0.8]$ (Ours)& \textbf{0.906} & \textbf{0.877} & \textbf{0.033} & \textbf{0.903} & \textbf{0.847} & \textbf{0.018} & \textbf{0.911} & \textbf{0.878} & \textbf{0.026}  \\
\bottomrule
\end{tabular}
}
\label{tab:ablation3}
\end{table}

3) Pooling Ratio in CGE: We design five variants to evaluate the impact of RGB node reduction ratios $r_{i}$ in the CGE graph pooling operation, including fixed ratios of ``$r_{i}=[0.2,0.2,0.2,0.2]$”, ``$r_{i}=[0.5,0.5,0.5,0.5]$”, and ``$r=[0.8,0.8,0.8,0.8]$”, a layer-wise decreasing ratio ``$r_{i}=[0.8,0.6,0.4,0.2]$”, and our top-down configuration ``$r_{i}=[0.2,0.4,0.6,0.8]$” as shown in Table~\ref{tab:ablation3}.
The fixed pooling strategies ``$r_{i}=[0.2,0.2,0.2,0.2]$”, ``$r_{i}=[0.5,0.5,0.5,0.5]$”, and ``$r_{i}=[0.8,0.8,0.8,0.8]$” reduce node redundancy to some extent; however, they fail to achieve optimal fusion performance due to their inability to adapt to different semantic levels. The layer-wise decreasing strategy ``$r_{i}=[0.8,0.6,0.4,0.2]$” provides better stability in retaining informative nodes, but its performance remains inferior to our top-down configuration, which adaptively balances compression strength across layers for improved semantic preservation and cross-scale alignment. From the results presented in Table~\ref{tab:ablation2}, we can see that although the fixed and layer-wise pooling strategies can reduce redundancy, they still do not effectively retain the most informative nodes at different semantic levels. These results emphasize the importance of a well-designed graph pooling strategy in suppressing noise while preserving discriminative and structurally meaningful information.

\begin{figure}[!htbp]
\centering
\includegraphics[width=0.49\textwidth]{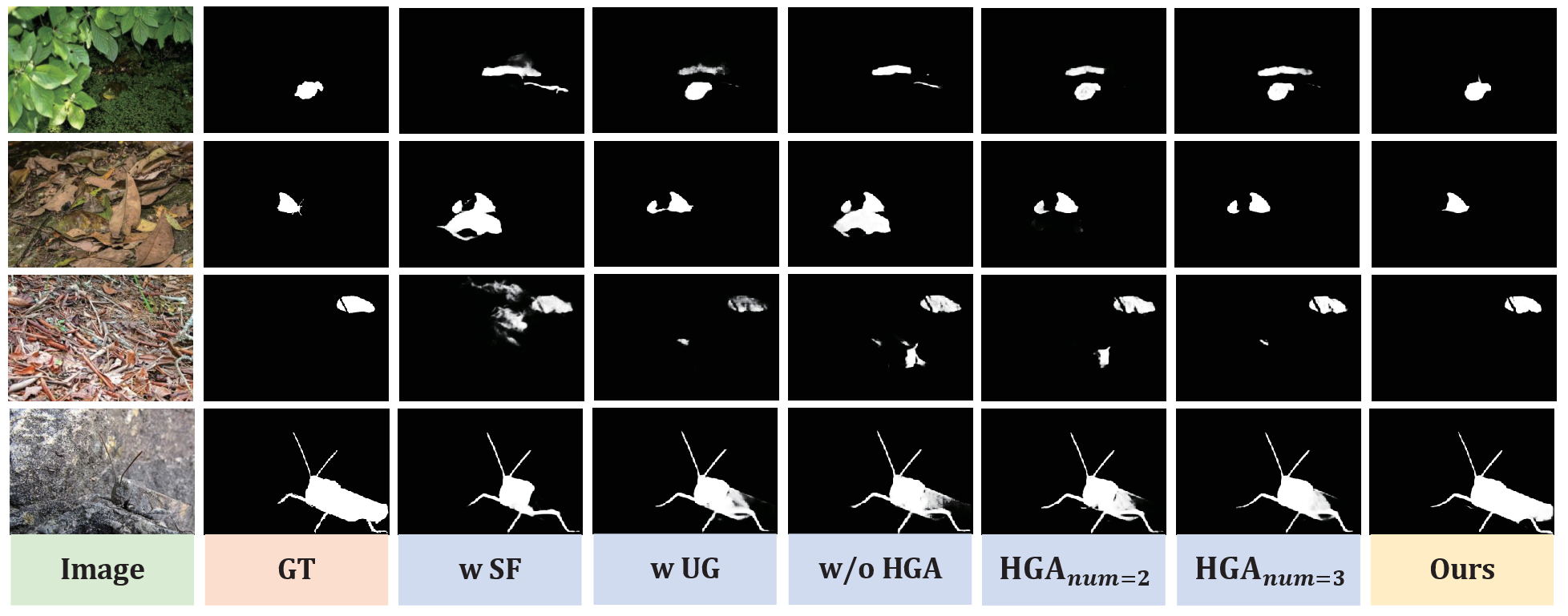} 
\caption{Visualization results of the ablation study of CGE.}
\label{fig:ab2}
\end{figure}

As shown in Fig.~\ref{fig:ab2}, the visual results of different CGE variants further validate the quantitative findings. The proposed heterogeneous graph fusion strategy generates more compact and complete target regions, effectively suppressing background noise while enhancing the precision of cross-modal feature alignment.

\subsubsection{Effectiveness of the AGR}
1) SSAG: To evaluate the effectiveness of the graph projection and semantic interaction stage, we design three variants: ``w/o SSAG", which completely removes the graph projection and selection branch, ``w/o GP\&UP", which disables the graph pooling and unpooling operations, and ``w/o Att", which removes the attention-based refinement in the fusion. As reported in Table~\ref{tab:ablation4}, comparing “w/o SSAG” with ``Ours" reveals that the SSAG effectively injects the global structural semantics from the PVT stream into the high-level SAM features. This enables the interaction process to produce more discriminative fused representations. Moreover, the performance drop of ``w/o GP\&UP" verifies the essential role of graph pooling and unpooling in the SSAG. These operations filter redundant or noisy nodes and perform hierarchical aggregation, thereby stabilizing feature propagation and improving the fusion result. Removing the attention mechanism (``w/o Att") leads to performance decline, indicating that attention-based refinement is effective in enhancing target representation. The results in Table~\ref{tab:ablation4} further demonstrate the effectiveness of the SSAG in facilitating interaction between the global structural semantics of the PVT features and the high-level SAM features, thereby significantly improving the localization and identification of camouflaged objects.

\begin{table}[h]
\renewcommand{\arraystretch}{1}
\renewcommand{\tabcolsep}{3.0mm}
\caption{Ablation study of the {\ModuleB~}}
\scalebox{0.65}{
\begin{tabular}{l|ccc|ccc|ccc}
\toprule
\multirow{2}{*}{\textbf{Method}} &\multicolumn{3}{c|}{\textbf{CAMO}}&\multicolumn{3}{c|}{\textbf{COD10K}}&\multicolumn{3}{c}{\textbf{NC4K}}\\ 
\cmidrule[0.05em](lr){2-10}
&$S_m\uparrow$&$F^{\omega}_{\beta}\uparrow$&$\mathcal{M}\downarrow$&$S_m\uparrow$&$F^{\omega}_{\beta}\uparrow$&$\mathcal{M}\downarrow$&$S_m\uparrow$&$F^{\omega}_{\beta}\uparrow$&$\mathcal{M}\downarrow$\\
\midrule
w/o SSAG & 0.885 & 0.840 & 0.043 & 0.876 & 0.789 & 0.023 & 0.899 & 0.847 & 0.032\\
w/o GP\&UP & 0.889 & 0.841 & 0.040 & 0.879 & 0.796 & 0.022 & 0.902 & 0.855 & 0.030\\
w/o Att & 0.895 & 0.853 & 0.039  & 0.885 & 0.810 & 0.021 & 0.905 & 0.863 & 0.029\\
\midrule
w/o CSP & 0.884 & 0.842 & 0.042 & 0.880 & 0.796 & 0.023 & 0.900 & 0.849 & 0.031\\
w/o Edge& 0.895 & 0.855 & 0.038 & 0.887 & 0.814 & 0.021 & 0.906 & 0.865 & 0.028\\
w Direct& 0.889 & 0.843& 0.041 & 0.884 & 0.811 & 0.021 & 0.902 & 0.857 & 0.030\\
\bottomrule
\rowcolor{row}Ours & \textbf{0.906} &\textbf{ 0.877} & \textbf{0.033} &\textbf{ 0.903} & \textbf{0.847} & \textbf{0.018} & \textbf{0.911} & \textbf{0.878} & \textbf{0.026} \\
\bottomrule
\end{tabular}
}
\label{tab:ablation4}
\end{table}

\begin{figure}[!htbp]
\centering
\includegraphics[width=0.47\textwidth]{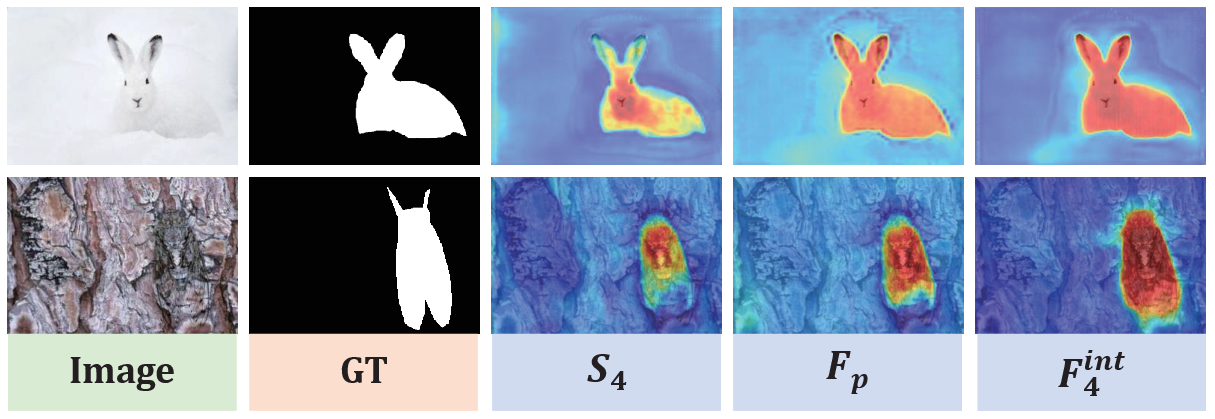} 
\caption{Visualization of feature maps for SSAG.}
\label{fig:hot1}
\end{figure}

To further verify the effectiveness of the SSAG branch, we visualize the intermediate feature maps in Fig.~\ref{fig:hot1}. It can be seen that the enhanced PVT feature $F_p$ provides clear structural cues of the camouflaged targets. Under the guidance of $F_p$, the SSAG branch progressively refines $S_4$, enabling it to better delineate object boundaries and suppress background noise. The final fused feature $F^{int}_4$ exhibits more compact and consistent activations within the target regions, demonstrating the importance of the proposed SSAG in accurately localizing the overall camouflaged object regions.

2) CSP: We conduct ablation experiments on the CSP by designing three variants: ``w/o CSP" completely removes the CSP, ``w/o Edge" disables the construction of graph edges and message passing, retaining only independent node-wise updates, and ``w Direct”, which bypasses the graph reasoning by directly injecting the SAM high-level feature $x_4$ into the branch without converting it into node representations, but instead fusing it with the input features through addition or concatenation. As shown in Table~\ref{tab:ablation4}, the removal of the entire CSP leads to a significant performance drop. This decline can be attributed to the absence of the progressive graph-based refinement process. The performance degradation of ``w/o Edge" further confirms the essential role of structural relation modeling between nodes within our AGR module. In addition, although ``w Direct" introduces $x_{4}$  as supplementary information, its performance still falls short of the full model, indicating that simple feature concatenation cannot replace the graph-based semantic refinement mechanism. 

\begin{figure}[!htbp]
\centering
\includegraphics[width=0.49\textwidth,height=0.27\textwidth]{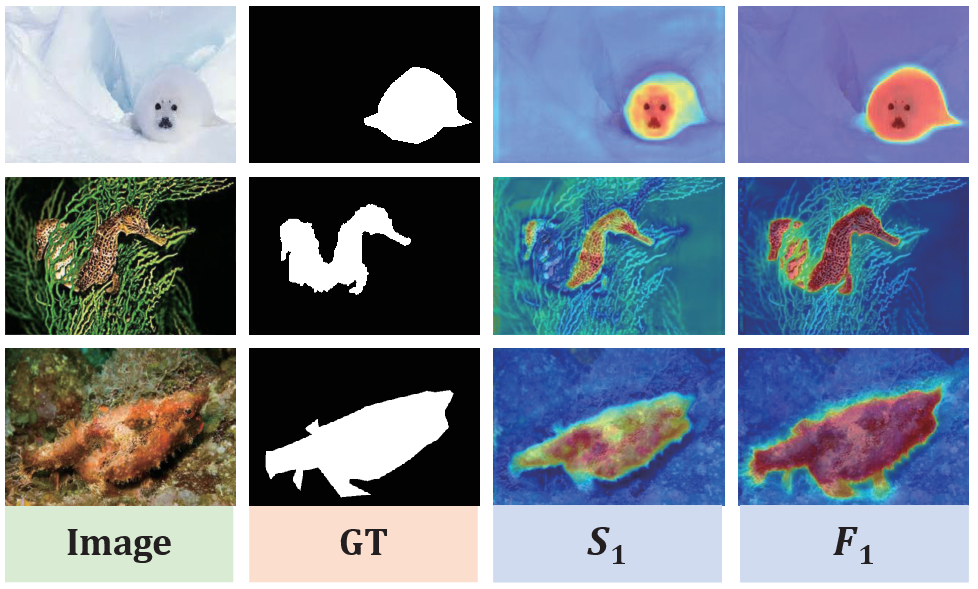}
\caption{Visualization of feature maps for CSP.}
\label{fig:hot2}
\end{figure}

3) Graph Pooling Ratio in SSAG: To further analyze the effect of the graph pooling ratio in the SSAG branch of AGR, we design four variants with different pooling ratios ``$r=0.3$, $r=0.5$, $r=0.9$". As shown in Table~\ref{tab:ablation5}, when $r$ is too small (e.g. ``$r=0.3$"), excessive node compression removes useful semantic information, resulting in incomplete feature propagation and degraded fusion quality. Increasing the ratio to ``$r=0.5$" alleviates this issue but still limits the representation capacity due to insufficient node diversity. Our adopted setting ``$r=0.7$" achieves the best trade-off between information retention and redundancy suppression, producing more stable graph reasoning and stronger semantic consistency. When the ratio further increases to ``$r=0.9$", redundant or noisy nodes are reintroduced, leading to over-smoothing and reduced discriminability. These results confirm that an appropriate graph pooling ratio is essential for achieving effective semantic aggregation and maintaining structural compactness in the SSAG process.

\begin{table}[h!]
\setlength{\tabcolsep}{4pt}
\caption{Ablation study on the graph pooling ratio in the SSAG branch of AGR.}
\centering
\scalebox{0.86}{
\begin{tabular}{c|ccc|ccc|ccc} 
   \toprule
    \multirow{2}*{$r$} &\multicolumn{3}{c|}{\textbf{CAMO}} &  \multicolumn{3}{c|}{\textbf{COD10K}}& \multicolumn{3}{c}{\textbf{NC4K}}\\
\cmidrule[0.05em](lr){2-10}
&$S_m\uparrow$&$F^{\omega}_{\beta}\uparrow$&$\mathcal{M}\downarrow$&
$S_m\uparrow$&$F^{\omega}_{\beta}\uparrow$&$\mathcal{M}\downarrow$&
$S_m\uparrow$&$F^{\omega}_{\beta}\uparrow$&$\mathcal{M}\downarrow$\\
\midrule
0.3 & 0.892 & 0.851 & 0.040 & 0.893 & 0.825 & 0.020 & 0.905 & 0.864 & 0.029\\
0.5 & 0.894 & 0.846 & 0.039 & 0.891 & 0.816 & 0.021 & 0.903 & 0.856 & 0.031\\
0.9 & 0.899 & 0.860 & 0.038 & 0.890 & 0.816 & 0.021 & 0.906 & 0.861 & 0.030\\
\midrule
\rowcolor{row}0.7 (Ours)& \textbf{0.906} & \textbf{0.877} & \textbf{0.033} & \textbf{0.903} & \textbf{0.847} & \textbf{0.018} & \textbf{0.911} & \textbf{0.878} & \textbf{0.026}  \\
\bottomrule
\end{tabular}
}
\label{tab:ablation5}
\end{table}

As shown in the visualization results in Fig.~\ref{fig:hot2}, the shallow SAM feature $S_1$ can roughly localize the target but lacks semantic completeness, while the refined feature $F_1$ exhibits more compact and coherent responses within the camouflaged object regions. This indicates that the CSP branch can effectively propagate the semantic information from the high-level feature $F^{int}_4$ to the shallow layers, achieving multi-scale refinement and enhancing the overall representation of camouf laged objects.

\begin{figure}[!htbp]
\centering
\includegraphics[width=0.48\textwidth]{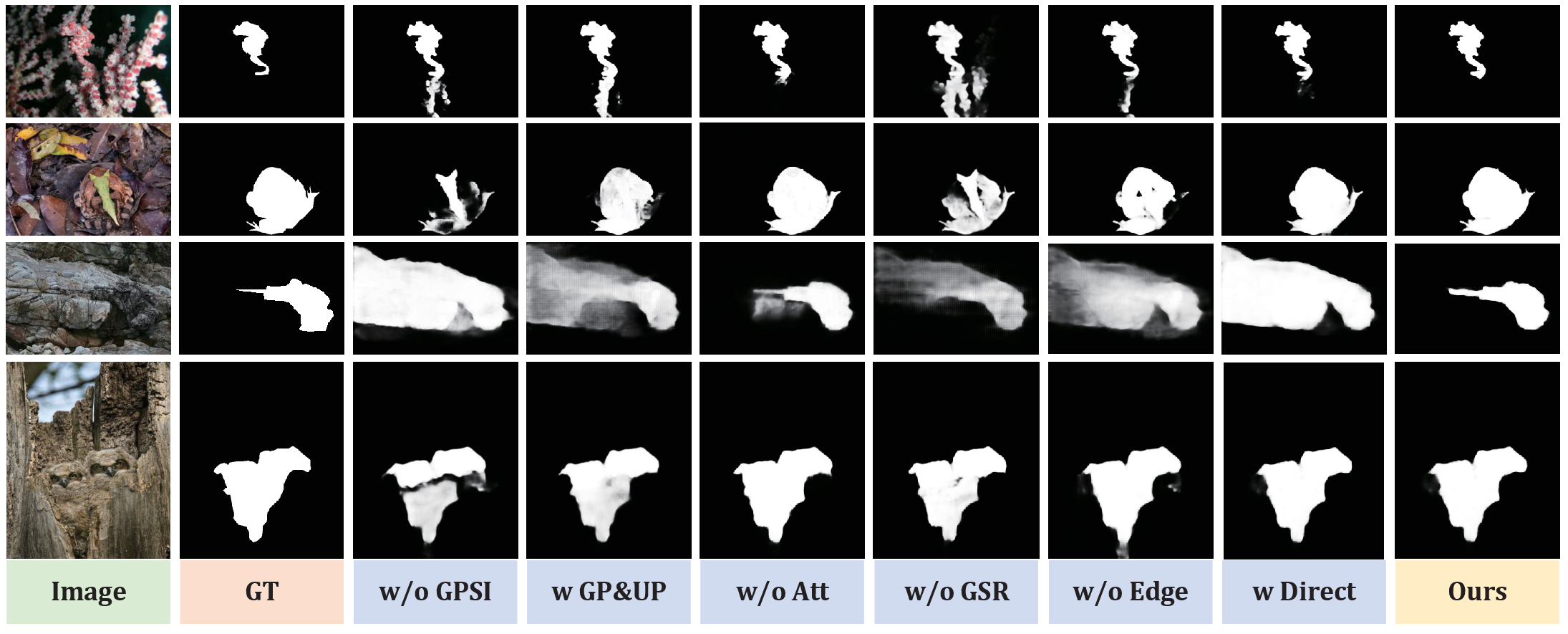} 
\caption{Visualization results of the ablation study of AGR.}
\label{fig:abAGR}
\end{figure}

As shown in Fig.~\ref{fig:abAGR}, the visual comparisons of different AGR variants further confirm the quantitative observations. The complete AGR module produces more compact and consistent activations with clearer object boundaries, demonstrating its effectiveness in enhancing cross-modal interaction and hierarchical semantic refinement.

\section{Conclution}
In this paper, we proposed DGA-Net, a SAM-based framework driven by depth prompting and graph modeling to detect camouflaged objects through geometric guidance, graph-based feature interaction, and semantic refinement. Specifically, we designed CGE to extract and align complementary information between RGB and depth modalities, thereby improving global structural awareness. In addition, AGR is introduced to refine SAM’s multi-level representations by injecting fine-grained semantics and preserving geometric consistency. Comprehensive experiments demonstrate that the proposed method outperforms state-of-the-art approaches. In our future work, we plan to explore the extension of our network to open-world and real-world segmentation scenarios, such as underwater exploration and industrial defect inspection in complex environments.



\bibliographystyle{elsarticle-num} 
\bibliography{reference}

%





\vfill

\end{document}